\documentclass{ws-procs11x85}
\usepackage{ws-procs-thm}           % comment this line when `amsthm / theorem / ntheorem` package is used
\usepackage{enumitem}
\usepackage{hyperref}
\usepackage{booktabs}

\begin{document}

\title{MedFactEval and MedAgentBrief: A Framework and Workflow for Generating and Evaluating Factual Clinical Summaries}

\author{
    François Grolleau, MD, PhD$^{1, \star}$,
    Emily Alsentzer, PhD$^2$,
    Timothy Keyes, PhD$^3$,\\
    Philip Chung, MD, MS$^4$,
    Akshay Swaminathan, BA$^2$,
    Asad Aali, MS$^5$,
    Jason Hom, MD$^7$, \\
    Tridu Huynh, MD$^7$, 
    Thomas Lew MD$^7$, 
    April Liang, MD$^7$, 
    Weihan Chu, MD$^7$, 
    Natasha Steele MD$^7$, \\
    Christina Lin MD$^7$, 
    Jingkun Yang  MD$^7$, 
    Kameron Black, MD$^7$, 
    Stephen Ma, MD, PhD$^{7}$, \\
    Fateme N. Haredasht, PhD$^1$,
    Nigam H. Shah, MBBS, PhD$^{1,2,7}$, 
    Kevin Schulman, MD, MBA$^{6,7}$,
    Jonathan H. Chen, MD, PhD$^{1,6,7}$
}

\address{
    $^1$Center for Biomedical Informatics Research, Stanford University, Stanford, CA, USA \\
    $^2$Department of Biomedical Data Science, Stanford University, Stanford, CA, USA \\
    $^3$Stanford Health Care, Palo Alto, CA, USA. \\
    $^4$Department of Anesthesiology and Pain Medicine, Stanford Medicine, Stanford, CA, USA \\
    $^5$Department of Radiology, Stanford University, Stanford, CA, USA \\
    $^6$Stanford Clinical Excellence Research Center, Stanford University, Stanford, CA, USA \\
    $^7$Department of Medicine, Stanford University, Stanford, CA, USA \\
    $^{\star}$To whom correspondence should be addressed: grolleau@stanford.edu
}

\begin{abstract}
Evaluating factual accuracy in Large Language Model (LLM)-generated clinical text is a critical barrier to adoption, as expert review is unscalable for the continuous quality assurance these systems require. We address this challenge with two complementary contributions. First, we introduce \textbf{MedFactEval}, a framework for scalable, fact-grounded evaluation where clinicians define high-salience key facts and an ``LLM Jury''—a multi-LLM majority vote—assesses their inclusion in generated summaries. Second, we present \textbf{MedAgentBrief}, a model-agnostic, multi-step workflow designed to generate high-quality, factual discharge summaries. To validate our evaluation framework, we established a gold-standard reference using a seven-physician majority vote on clinician-defined key facts from inpatient cases. The MedFactEval LLM Jury achieved almost perfect agreement with this panel (Cohen's $\kappa=81\%$), a performance statistically non-inferior to that of a single human expert ($\kappa=67\%, P < 0.001$). Our work provides both a robust evaluation framework (MedFactEval) and a high-performing generation workflow (MedAgentBrief), offering a comprehensive approach to advance the responsible deployment of generative AI in clinical workflows.
\end{abstract}

\section{Introduction}
Large Language Models (LLMs) offer substantial potential for automating clinical documentation, a primary driver of physician burnout \cite{moy2021measurement}. By generating draft discharge summaries, LLMs can perform critical information synthesis, freeing time for clinicians to focus on patient care \cite{van2024adapted, williams2025physician}. However, the risk of factual errors impedes their adoption \cite{mesko2023imperative}. These errors manifest as both errors of \textit{omission}, where clinically vital information is excluded, and errors of \textit{commission} (or ``hallucinations"), where generated text contradicts source data or fabricates information \cite{morgan2025time, hong2025application}.

The gold standard for identifying such errors—expert human review—is prohibitively slow and expensive for the routine, iterative quality assurance these systems require. This creates a critical evaluation bottleneck, as current automated metrics (e.g., BLEU, ROUGE-L, BERT-Score) often fail to capture clinical nuance. Consequently, there is an urgent need for an evaluation framework that is both scalable and replicates the clinical validity of expert judgment.

To address this challenge, this paper makes two complementary contributions. First, we introduce and validate \textbf{MedFactEval}, a framework designed to enable the fact-grounded, iterative improvement of AI systems for clinical information synthesis. Second, we introduce \textbf{MedAgentBrief}, a novel, model-agnostic LLM workflow for generating high-quality, factual discharge summaries. We then demonstrate the utility of MedFactEval by using it to benchmark the performance-cost tradeoffs of these different generation strategies.

Our specific contributions are:
\begin{enumerate}[label=\arabic*., itemsep=0pt, topsep=3pt]
    \item We define the \textbf{MedFactEval} framework, which grounds evaluation in clinician-defined key facts and uses an LLM Jury for robust, scalable assessment to drive iterative model improvement.
    \item We introduce and evaluate \textbf{MedAgentBrief}, a multi-step, model-agnostic workflow for generating high-fidelity clinical summaries.
    \item We conduct a rigorous meta-evaluation of MedFactEval, demonstrating that our LLM Jury achieves almost perfect agreement with a seven-physician panel (Cohen's $\kappa = 81\%$) and is statistically non-inferior to a single human expert ($\kappa=67\%, P < 0.001$).
    \item We demonstrate MedFactEval's utility by benchmarking the factual performance of MedAgentBrief against a baseline single-prompt strategy.
\end{enumerate}
Our primary contribution is the validation of the MedFactEval framework, establishing that an LLM Jury can serve as a reliable and scalable proxy for expert human evaluation. We then demonstrate the framework's utility by applying it to benchmark generation strategies, providing a foundational component for the safety monitoring of generative AI in clinical medicine.

\section{Related Work}
The evaluation of LLM-generated medical text is a rapidly evolving field. Early approaches relied on comparing AI-generated summaries against human-written references using global quality ratings from physician experts \cite{van2024adapted,williams2025physician,asgari2025framework}. While these foundational studies established that LLMs can produce summaries of comparable quality to humans, they also highlighted a persistent risk of factual errors, including both omissions and commissions. However, such global ratings are inherently subjective and often demonstrate low inter-rater reliability. This makes them a noisy signal, insufficient for the granular, fact-based feedback required to systematically improve model factuality and ensure patient safety.

To address the need for more objective evaluation, automated fact-checking has emerged as a key area of research\cite{zuo2024medhallbench, chung2025verifact}. For instance, ``VeriFact" is a system that retrospectively verifies claims in generated text by decomposing them into atomic facts and checking them against the entire electronic health record (EHR) using retrieval-augmented generation.\cite{chung2025verifact} While powerful, this approach faces challenges: the EHR itself can be noisy or contain outdated information, and the system does not inherently know which facts are most clinically salient for a specific care context. Other work has confirmed that LLMs can struggle with fact decomposition, for example, the FactEHR effort showed that different LLMs generate up to 2.6x different number of facts per sentence \cite{munnangi2024assessing}. In addition, standard natural language processing metrics often fail to correlate with human judgments of factuality \cite{tang2023evaluating}. 

Our MedFactEval framework is distinct from prior work in three key ways. First, it is designed for \textbf{ongoing evaluation of an in-use workflow} rather than for one-off, retrospective analysis, providing a mechanism for continuous quality assurance. Second, it shifts the source of ground truth from the entire, potentially noisy EHR to a small set of \textbf{clinician-defined, key facts}. This strategically reframes the ambiguous task of ``is this summary good?" into a series of concrete, verifiable questions (e.g., ``Does the summary include Fact X?"). This ensures the evaluation is focused on what matters most for a given patient context. Finally, it introduces an \textbf{LLM Jury} to automate this assessment, enhancing robustness and reliability over single-model judgments. \cite{zheng2023judging} These design choices position MedFactEval as a pragmatic tool for both real-time quality assurance and the iterative improvement of AI systems in clinical practice.

\section{Methods}
\subsection{Study Setting and Data Source}
We conducted a retrospective study using a cohort of adult inpatients to validate MedFactEval, our framework for the ongoing evaluation of in-use clinical AI workflows. The cohort was drawn from patients admitted to the Division of Hospital Medicine at Stanford Health Care between January 1, 2023, and June 1, 2024. To ensure representative cases of sufficient clinical complexity, we included only patients with a hospital length of stay between 2 and 14 days. From this eligible cohort, we randomly sampled 30 patients. For each patient, we extracted the complete set of de-identified, unstructured clinical notes from the Electronic Health Record (EHR), including the History \& Physical, all progress notes, and the physician-authored discharge summary.

\subsection{Generation of AI Discharge Summaries for Evaluation}
For each of the 30 cases, the collected clinical notes served as source material to generate a diverse set of draft discharge summaries. We employed two distinct AI synthesis strategies to produce summaries for evaluation.

\paragraph{Strategy 1: Single-Prompt Summarization.}
In this baseline approach, all source notes for a given patient were concatenated and provided within the context window of a foundation model. The model was then prompted to generate a complete discharge summary in a single step.

\paragraph{Strategy 2: MedAgentBrief Workflow.}
We developed MedAgentBrief, a model-agnostic, multi-step workflow designed to enhance factual accuracy through an iterative process:

\begin{enumerate}[label=(\roman*), itemsep=0pt, topsep=2pt]
    \item \textbf{Initial Draft Generation:} A foundation model produced a first-pass summary using only the History \& Physical and the final progress note as source documents.
    \item \textbf{Iterative Refinement:} The initial draft was then iteratively refined by processing the remaining progress notes in chronological order. For each note, a prompt instructed the model to identify salient medical information, integrate it into the evolving summary, and simultaneously embed a corresponding provenance tag (e.g., \texttt{<PROG\_NOTE\_7>}) with each new statement.
    \item \textbf{Hallucination Reduction \& Citation:} A concluding step first performed a verification pass, reviewing the summary to identify and mitigate unsupported factual claims. Following this check, the embedded provenance tags were resolved, linking each tag to its full source note to enable downstream citation generation.
\end{enumerate}
An anonymized sample of a MedAgentBrief output is available as \href{https://fcgrolleau.github.io/medfacteval/samples/anonymized_medagentbrief_sumary.html}{Supplementary Data 1}.

\paragraph{Foundation Models and Output Structure.}
These two strategies were implemented using various foundation models, such as GPT-4o and DeepSeek-R1 (see Supplementary Table S1 for a complete list), all accessed via a secure, HIPAA-compliant Application Programming Interface (API) infrastructure to ensure data privacy \cite{ng2025development}. The scope of our generation task focused on the narrative synthesis of a patient's stay. All models were prompted to produce a consistent, structured output containing: (1) a one-line summary, (2) a brief narrative of the hospital course, and (3) a problem-focused summary. This task was explicitly scoped to synthesizing information already present in the source notes. The generation of boilerplate text (e.g., standard patient instructions) and novel medical recommendations not found in the provided documentation was considered out of scope.

\vspace{.3cm}\subsection{The MedFactEval Framework}
The MedFactEval framework consists of three main steps: benchmark creation, LLM Jury evaluation, and automated reporting.
\paragraph{Step 1: Benchmark Creation---Key Fact Extraction.}
For each patient case, one or more senior attending physicians reviewed the original, human-written discharge summary to identify three high-salience clinical facts. These ``key facts" represent critical information that a high-quality summary must not omit. Examples of such key facts include: (i) a new critical diagnosis, (ii) a key change in management, or (iii) an essential follow-up action.

To streamline this process, annotators were optionally provided with candidate key facts suggested by a Task-Assisting LLM (based on models such as GPT-4o and Gemini 2.0 Flash), which they could then review, edit, or discard. This process yields a benchmark of clinician-validated facts, creating a trusted reference for evaluating AI summaries on their ability to capture clinically critical information.

\paragraph{Step 2: LLM Jury Evaluation.}
To evaluate a generated summary, our framework employs an ``LLM Jury" of ten distinct LLM judges. For each key fact, the jury is prompted to assess whether the summary includes that fact. Each judge provides a binary verdict and a brief justification, with the final decision determined by a majority vote. A parallel process is used to assess for contradictions, where the jury determines if the summary contains information inconsistent with the key fact.

\paragraph{Step 3: Automated Evaluation Report.}
To synthesize the evaluation results into an actionable format, the framework generates an automated report for each evaluated AI system. This report uses a language model to distill qualitative themes from the LLM Jury's explanations for omitted or contradicted facts. The final output programmatically aggregates these insights with quantitative performance scores into a structured HTML document.  An anonymized sample of a complete HTML report is available as \href{https://fcgrolleau.github.io/medfacteval/samples/anonymized_medfacteval_report.html}{Supplementary Data 2}.

\subsection{Meta-Evaluation of the MedFactEval Framework}
The validity of MedFactEval hinges on whether its automated scores correspond with expert human judgment. We therefore conducted a meta-evaluation to assess this correspondence.

\paragraph{Gold Standard Creation.}
We created a gold standard reference using a subset of the generated summaries. Specifically, summaries generated for all 30 cases by two MedAgentBrief systems (one using GPT-4o, the other DeepSeek-R1) were selected, resulting in 60 unique summaries for evaluation. A panel of seven attending physicians independently evaluated each summary, providing a binary judgment (present/absent) for each of the three corresponding key facts. The gold standard for fact presence was then defined as the majority vote of this seven-physician panel.

\paragraph{Human Expert Baseline.}
To establish a performance baseline for a single human expert, we calculated the agreement for each of the seven physicians against the majority vote of the other six (a leave-one-out approach). The final human expert baseline is then reported as the average of these seven individual agreement scores.

\paragraph{Statistical Analysis.}
Our primary outcome was the inter-rater agreement between an evaluator (MedFactEval LLM Jury or a single human expert) and the seven-physician gold standard. We measured agreement using Cohen's kappa coefficient ($\kappa$) and used a one-sided test to assess for non-inferiority against a pre-specified margin of $10\%$ for the kappa coefficient\cite{landis1977measurement}. Ninety-five percent confidence intervals (95\% CI) were calculated by bootstrapping (9999 iterations).

\section{Results}

\subsection{An Instantiated Benchmark for Discharge Summaries}
Applying the first step of the MedFactEval framework—Key Fact Extraction—we developed a benchmark for the task of discharge summary generation. This benchmark comprises three core components: First, it specifies the \textbf{task} as the synthesis of a patient's hospital course into a structured discharge summary. Second, it provides the \textbf{source data}: the complete set of unstructured clinical notes for each of 30 patients, including the History \& Physical and all progress notes from both physician (attending and consultant) and allied health services (e.g., clinical nutrition, occupational therapy, speech language pathology). Finally, it establishes the \textbf{evaluation standard}: a set of 90 clinician-defined ``key facts" (three per case), which provide the concrete criteria for assessing the factual integrity of a generated summary.

Examples of such key facts include new critical diagnoses (e.g., ``admitted for recurrent malignant biliary obstruction with hyperbilirubinemia requiring percutaneous transhepatic drainage"), key changes in management (e.g., ``discharged with deep vein thrombosis prophylaxis, with anemia at hemoglobin 9g/dL"), and essential follow-up actions (e.g., ``will return next week to have the small skin cancer on the right cheek removed"). The specific characteristics of this benchmark are detailed in Table \ref{table:benchmark_chars}.

\begin{table}[h!]
\centering
\tbl{\textbf{Benchmark Characteristics.} SD: Standard Deviation, IQR: InterQuartile Range.\label{table:benchmark_chars}}
{\begin{tabular}{@{}lc@{}}
    \toprule
    \textbf{Characteristic} & \textbf{Value} \\
    \midrule
    \multicolumn{2}{l}{\textit{Patient Demographics}} \\
    \hspace{1em} Number of patients & 30 \\
    \hspace{1em} Age, mean (SD), years & 60 (16) \\
    \hspace{1em} Sex, n (\%) & \\
    \hspace{2em} Female & 17 (57) \\
    \hspace{2em} Male & 13 (43) \\
    \hspace{1em} Length of stay, median (IQR), days & 5 (4--8) \\
    \midrule
    \multicolumn{2}{l}{\textit{Source Documents (per case)}} \\
    \hspace{1em} Total progress notes, median (IQR) & 12 (10--16) \\
    \hspace{1em} Physician notes, median (IQR) & 9 (7--11) \\
    \hspace{1em} Allied health notes, median (IQR) & 3 (2--6) \\
    \midrule
    \multicolumn{2}{l}{\textit{Key Fact Categorization}} \\
    \hspace{1em} Total Key Facts & 90 \\
    \hspace{1em} Diagnosis-related, n (\%) & 46 (51) \\
    \hspace{1em} Management Change, n (\%) & 36 (40) \\
    \hspace{1em} Follow-up Action, n (\%) & 5 (6) \\
    \hspace{1em} Other, n (\%) & 3 (3) \\
    \bottomrule
\end{tabular}}
\end{table}

\begin{figure}[h!]
    \vspace{0pt}  % force top alignment
    \centering
    \includegraphics[width=\linewidth]{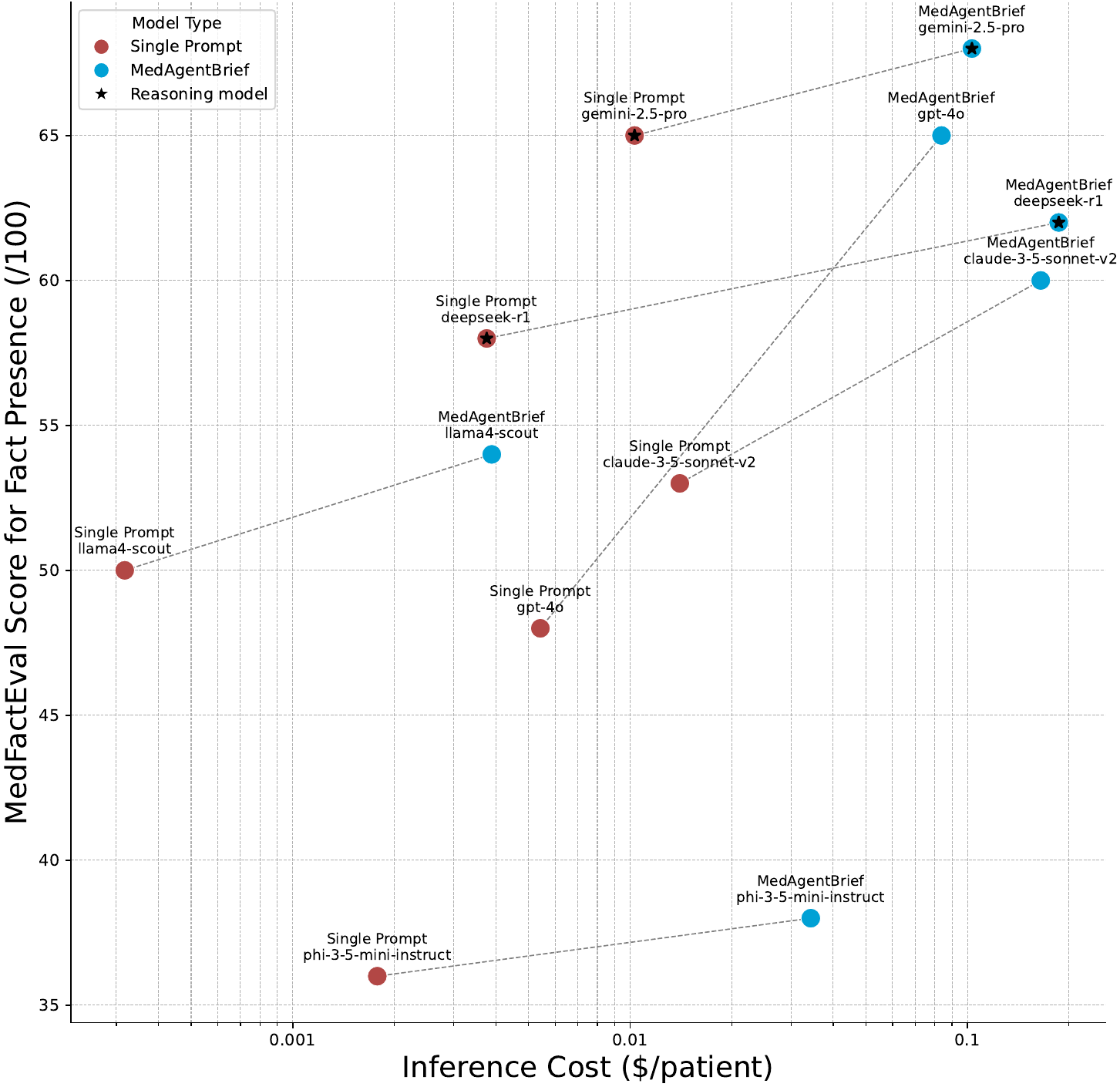}
    \caption{
    \textbf{Performance-Cost Trade-off of AI Summary Generation, Measured by MedFactEval.} 
    The y-axis represents the MedFactEval score for factual presence (higher is better), while the x-axis shows inference cost per patient on a logarithmic scale. For each foundation model, the MedAgentBrief workflow (cyan) consistently yields a higher factuality score than the baseline Single Prompt approach (brown), demonstrating its effectiveness at the expense of increased cost. Models designated as ``Reasoning models" (starred) generally occupy the high-performance, high-cost quadrant.
}\label{fig:bench_res}
\end{figure}

\subsection{Benchmarking Generation Strategies with MedFactEval}
We applied MedFactEval to compare our two generation strategies—the baseline Single-Prompt approach and the MedAgentBrief workflow—across various foundation models. 

\paragraph{Fact Presence vs. Inference Cost.}
As illustrated in Figure \ref{fig:bench_res}, the MedAgentBrief workflow consistently improved factual presence scores over the Single-Prompt baseline for every foundation model tested. This performance gain, however, came at an increased inference cost, highlighting a clear trade-off between factuality and expense. A parallel trade-off between performance and generation time (latency) was also observed (see Supplementary Figure S1). While the models designated as ``Reasoning models" in our benchmark (e.g., DeepSeek-R1, Gemini-2.5-Pro) occupied the high-performance quadrant, other powerful models also benefited significantly. For instance, the factual presence score of GPT-4o improved by 17 percentage points (from 48\% to 65\%) when using the MedAgentBrief workflow.

\paragraph{Analysis of Factual Errors.}
MedFactEval identified contradiction rates ranging from 10\% to 25\% across the different systems, with performance varying by both foundation model and generation strategy (see Supplementary Figure S2). A manual review revealed that this rate is attributable to MedFactEval LLM Jury's high sensitivity in detecting subtle yet clinically significant inconsistencies. For instance, the jury correctly flagged a summary that misrepresented a complex fondaparinux-to-apixaban transition as a simple initiation. This distinction is critical, as it obscures the patient's prior anticoagulation history and the clinical reasoning for the change (e.g., treatment failure or convenience), information vital for future care decisions. While notable, these contradiction rates were consistently lower than errors of omission, making fact presence the primary differentiator between systems.

\subsection{Meta-Evaluation: Validating MedFactEval LLM Jury}
\begin{figure}[h!]
    \centering
    \includegraphics[width=\linewidth]{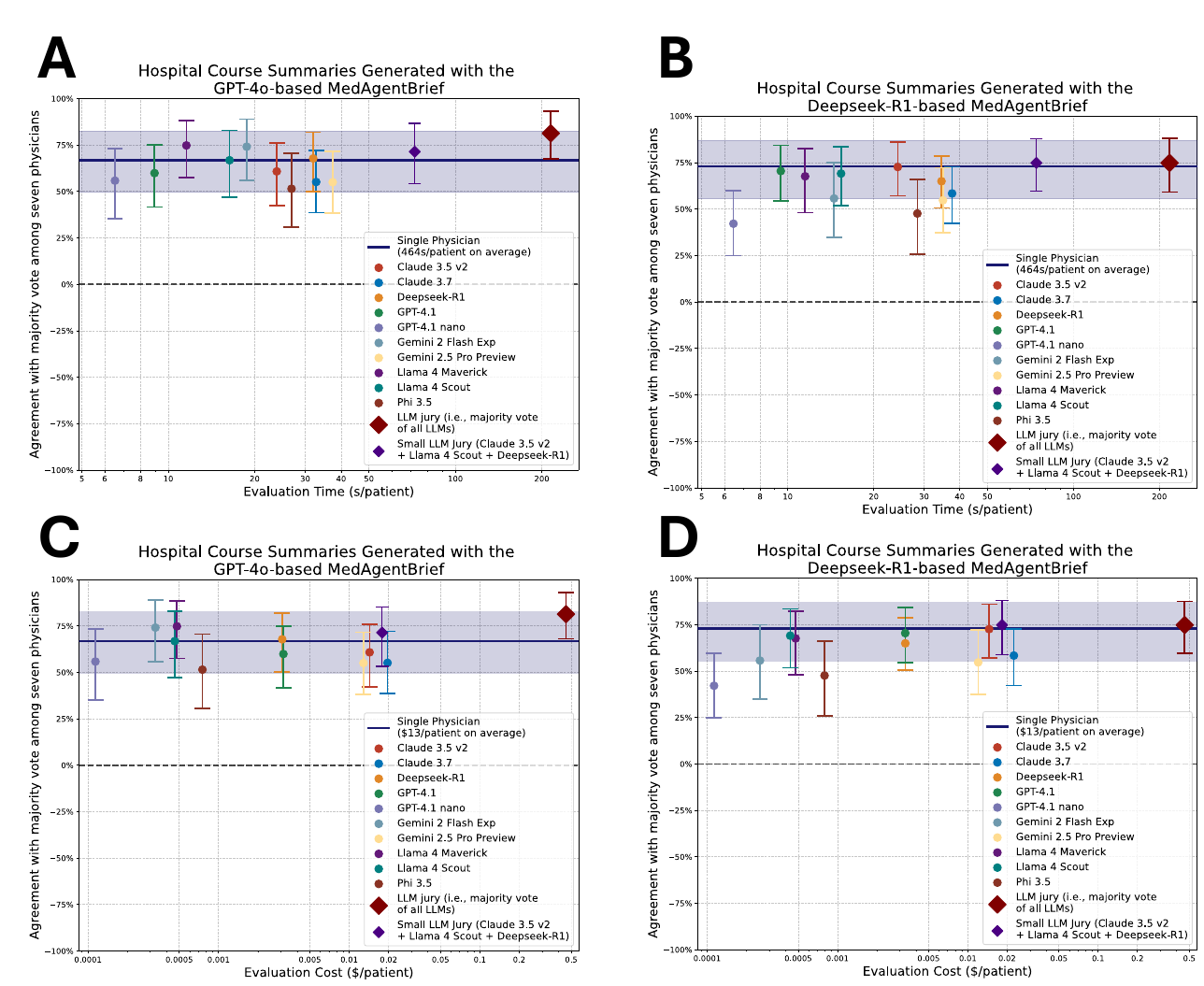}
\caption{
    \textbf{Agreement of LLM Judges and Juries with the Physician Gold Standard.} 
    Each point represents the Cohen's kappa agreement for an evaluator against the seven-physician majority vote. 
    \textbf{(A, C)} Agreement versus evaluation time for summaries from the GPT-4o and Deepseek-R1 workflows, respectively. 
    \textbf{(B, D)} Agreement versus evaluation cost for the same summaries. 
    The full 10-member LLM Jury (red diamond) consistently achieves higher agreement than most individual judges. The Small LLM Jury (purple diamond) offers a faster, lower-cost alternative while maintaining performance comparable to the single physician reference (blue shaded area, representing the 95\% CI for a single physician's agreement).
}\label{fig:metaeval}
\end{figure}
Our primary goal was to validate whether the MedFactEval LLM Jury could serve as a reliable proxy for human expert evaluation.

\paragraph{LLM Jury Agreement with Gold Standard.}
As shown in Figure \ref{fig:metaeval}, the 10-member LLM Jury demonstrated almost perfect agreement with the seven-physician majority vote gold standard for determining fact presence. This strong performance was observed across both systems evaluated: for summaries from the GPT-4o-based MedAgentBrief, the jury achieved a Cohen's kappa of $\kappa = 81\%$ (95\% CI: 66\%--92\%), while for the DeepSeek R1-based MedAgentBrief system, the agreement was $\kappa = 75\%$ (95\% CI: 59\%--88\%). This level of agreement was substantially higher than that achieved by any individual LLM judges, underscoring the value of the jury ensemble approach.

\begin{figure}[h!]
    \centering
    \includegraphics[width=\linewidth]{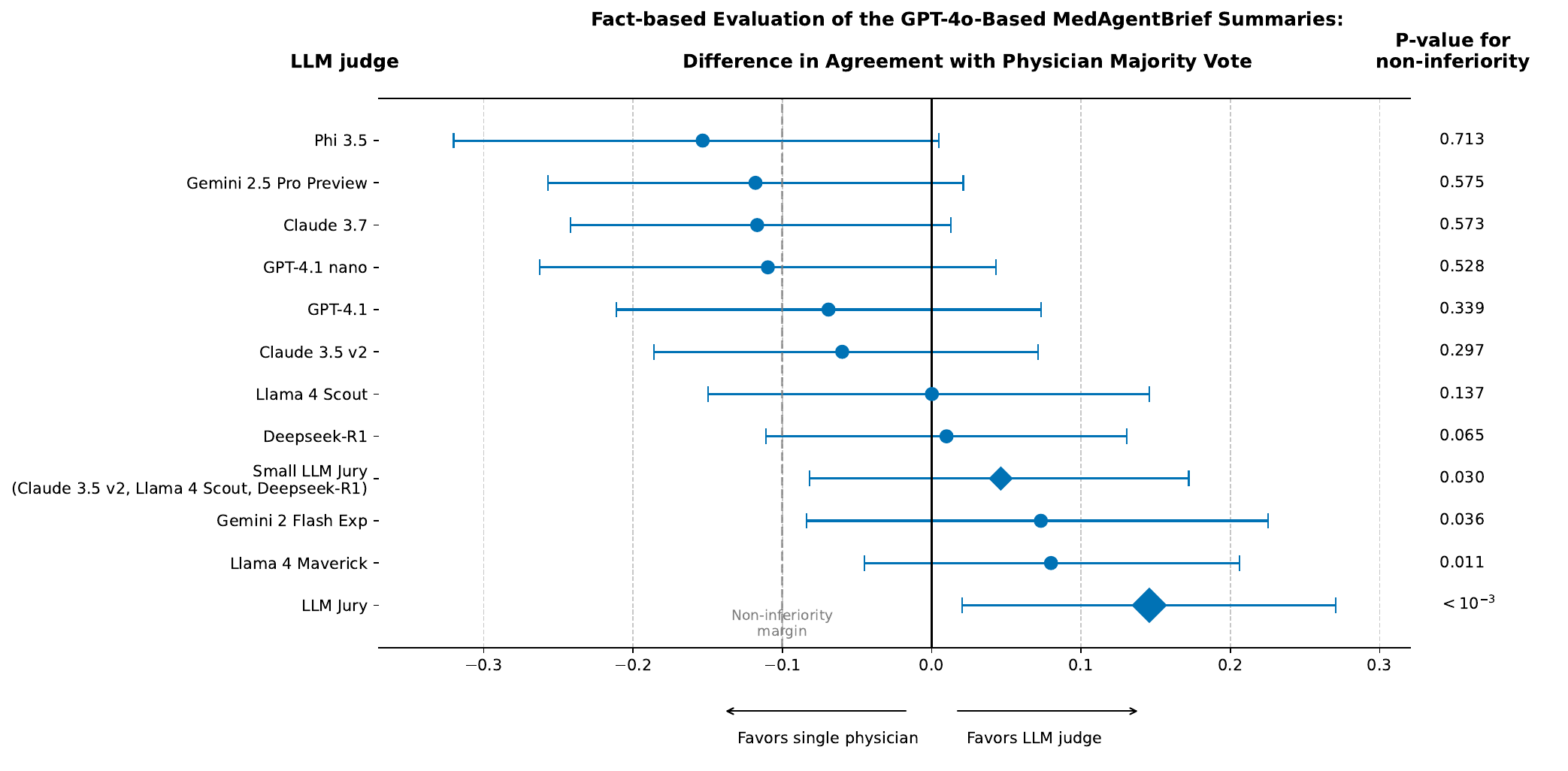}
    \caption{\textbf{Non-Inferiority of LLM Judges Compared to a Single Physician Expert.} The plot shows the difference in Cohen's kappa between each LLM judge and the average single physician, for summaries generated by the GPT-4o-based MedAgentBrief. Points to the right of the vertical line favor the LLM judge. The dashed line indicates the pre-specified non-inferiority margin. The full LLM Jury (bottom diamond) and the Small LLM Jury both surpassed the single physician baseline and met the criteria for non-inferiority ($P < 0.001$). Error bars represent 90\% confidence intervals, consistent with the one-sided nature of the non-inferiority test.}\label{fig:noninf_gpt4o}
\end{figure}

\paragraph{Non-Inferiority to a Single Human Expert.}
Our analysis confirmed that the MedFactEval LLM Jury is statistically non-inferior to a single human expert for evaluating factual presence (Figure \ref{fig:noninf_gpt4o}). When evaluating summaries from the GPT-4o-based MedAgentBrief, the full LLM Jury's agreement ($\kappa = 81\%$) was substantially higher than the single-physician baseline ($\kappa = 67\%$, 95\% CI: 49\%--83\%). This resulted in a kappa difference of 15\% (95\% CI: -1\%--29\%), satisfying our non-inferiority criteria ($P < 0.001$). This finding of non-inferiority also held when evaluating the Deepseek R1-based system, which showed a kappa difference of 2\% (95\% CI: -11\%--15\%; $P < 0.001$) (see Supplementary Figure S3). Furthermore, as shown in Figure \ref{fig:noninf_gpt4o}, the cost-effective ``Small LLM Jury" (comprising Claude 3.5, Llama-4-Scout, and DeepSeek-R1) also met the non-inferiority threshold, suggesting that practical, lower-cost versions of the jury are viable for real-world application.

\section{Discussion}
In this work, we introduced and validated MedFactEval, a scalable framework for fact-grounded evaluation, and presented MedAgentBrief, a high-performing workflow for generating clinical summaries. Our primary finding is that an LLM Jury, guided by clinician-defined key facts, can evaluate the factual accuracy of AI-generated text with a reliability non-inferior to that of a single human physician expert. While we focused on discharge summaries, the MedFactEval methodology is intentionally generalizable and can be applied to other clinical information synthesis tasks by defining a new set of relevant key facts.

This result is significant because it offers a path to overcoming the critical evaluation bottleneck in clinical AI. Manual, expert-led review is unsustainable for the continuous monitoring required for safe deployment. By demonstrating that an LLM Jury can serve as a reliable and scalable proxy, MedFactEval provides a practical tool for real-time quality assurance and iterative model improvement. This moves evaluation from a retrospective academic exercise toward a dynamic, automated process suitable for live clinical environments.

Furthermore, our benchmarking results highlight the utility of MedFactEval for concrete development tasks. The clear performance-cost trade-off revealed in Figure \ref{fig:bench_res} can guide decisions about model and strategy selection. The framework's ability to systematically identify errors of omission—which our analysis confirmed is the predominant failure mode—provides developers with the granular, actionable feedback needed to improve the safety and reliability of their systems.

\subsection*{Limitations and Future Directions}
Our study has several limitations. The validation was conducted on a cohort from a single service-line (general inpatient medicine) from a single academic medical center. While we demonstrated that even a small, locally-created benchmark can provide a powerful signal for model selection and improvement, larger-scale validation across diverse settings is warranted. Second, our framework intentionally focuses on a small number of clinician-defined, high-salience facts (three per case in our study), a constraint chosen to mirror the low cognitive burden required for practical integration into a busy clinical workflow. This ensures clinical relevance but does not assess other quality dimensions like fluency or conciseness, nor does it detect the omission of facts not pre-specified by the clinician. MedFactEval is therefore intended as a targeted fact-checking tool to be used alongside other holistic evaluation methods. Integrating our framework into comprehensive platforms like MedHELM\cite{bedi2025medhelm} would support this goal by enabling multi-dimensional assessment. Finally, the process of selecting key facts is inherently subjective. This was a deliberate design choice, meant to empower physicians to direct the evaluation toward what they deem clinically critical. However, future work could explore methods to further standardize or automate this selection process.

Looking forward, our immediate goal is to deploy MedAgentBrief in a pilot study, using MedFactEval to provide real-time safety monitoring. In this prospective workflow, a clinician would define a small number of critical facts for their patient's summary. The MedFactEval jury would then assess the AI-generated draft against these facts, providing immediate feedback on whether the facts were correctly included or contradicted. This real-time loop would empower clinicians to identify AI failure modes as they occur. Furthermore, this process would generate a prospectively collected, real-world dataset of clinical notes paired with expert-defined key facts, providing an invaluable resource for future model training and evaluation.

\section{Conclusion}
This work advances the safe and effective use of AI in clinical documentation by providing two key contributions: a robust evaluation framework, MedFactEval, and a high-performing generation workflow, MedAgentBrief. By demonstrating that an LLM Jury can reliably and scalably replicate expert judgment for fact-checking, we offer a practical solution for the continuous safety monitoring required for the responsible deployment of AI systems for clinical information synthesis.

\section*{Supplementary Materials}
Supplementary materials associated with this article can be found at:\\
\href{https://fcgrolleau.github.io/medfacteval/supplementary_materials.pdf}{https://fcgrolleau.github.io/medfacteval/supplementary\_materials.pdf}\\
This includes Supplementary Table S1 and Supplementary Figures S1-S3.

\section*{Code Availability}
The complete source code for the MedFactEval framework, the MedAgentBrief workflow, and all experiments presented in this paper is open-source and publicly available on GitHub at: \href{https://github.com/fcgrolleau/medfacteval}{https://github.com/fcgrolleau/medfacteval}.

\section*{Data Availability}
The validation benchmark used in this study contains detailed clinical notes of recently admitted patients at Stanford Health Care. Due to the sensitive nature of this data and the risk of direct re-identification, it cannot be publicly shared. The following supplementary data are available with this article:
\begin{itemize}[itemsep=0pt, topsep=2pt, leftmargin=15pt]
    \item \textbf{Supplementary Data 1:} Anonymized sample of a MedAgentBrief-generated summary (HTML format). Available at:\\
    \href{https://fcgrolleau.github.io/medfacteval/samples/anonymized_medagentbrief_sumary.html}{https://fcgrolleau.github.io/medfacteval/samples/anonymized\_medagentbrief\_sumary.html}

    \item \textbf{Supplementary Data 2:} Anonymized sample of an automated MedFactEval evaluation report (HTML format). Available at:\\
    \href{https://fcgrolleau.github.io/medfacteval/samples/anonymized_medfacteval_report.html}{https://fcgrolleau.github.io/medfacteval/samples/anonymized\_medfacteval\_report.html}
\end{itemize}

\clearpage

\bibliographystyle{ws-procs11x85}
\bibliography{ws-pro-sample}

\end{document}